\pdfoutput=1

\documentclass[11pt]{article}

\usepackage{acl}

\usepackage{times}
\usepackage{latexsym}

\usepackage[T1]{fontenc}

\usepackage[utf8]{inputenc}

\usepackage{microtype}

\usepackage{soul} %
\usepackage{graphicx}
\usepackage{bbding}
\usepackage{booktabs}
\usepackage{multirow}
\usepackage{tablefootnote}
\usepackage{wrapfig}

\usepackage{tikz}
\usepackage{tikz}
\newcommand*\circled[1]{\tikz[baseline=(char.base)]{
            \node[shape=circle,draw,inner sep=1.5pt] (char) {#1};}}

\usepackage{tabularx}
\usepackage{makecell}
\newcolumntype{R}[1]{>{\hsize=#1\hsize\raggedleft\arraybackslash}X}%
\newcolumntype{L}[1]{>{\hsize=#1\hsize\raggedright\arraybackslash}X}%
\newcolumntype{C}[1]{>{\hsize=#1\hsize\centering\arraybackslash}X}%

\title{LM-CORE:
\underline{L}anguage \underline{M}odels with \underline{Co}ntextually \underline{R}elevant \underline{E}xternal Knowledge}

\author{Jivat Neet Kaur\Thanks{ Work done as an intern at the Media and Data Science Research Lab, Adobe, India.}$\,\;^\alpha$, Sumit Bhatia$^\beta$, Milan Aggarwal$^\beta$,\\ \textbf{Rachit Bansal$^{*\gamma}$}, \and \textbf{Balaji Krishnamurthy$^\beta$} \\
$^\alpha$ Microsoft Research, India \\
$^\beta$ Media and Data Science Research Lab, Adobe, India \\
$^{\gamma}$ Delhi Technological University, India \\
\texttt{t-kaurjivat@microsoft.com, sumit.bhatia@adobe.com} \\
\texttt{milaggar@adobe.com, racbansa@gmail.com, kbalaji@adobe.com}
}

\begin{document}

\maketitle

\begin{abstract}
Large transformer-based pre-trained language models have achieved impressive performance on a variety of knowledge-intensive tasks and can capture factual knowledge in their parameters. We argue that storing large amounts of knowledge in the model parameters is sub-optimal given the ever-growing amounts of knowledge and resource requirements. We posit that a more efficient alternative is to provide explicit access to contextually relevant structured knowledge to the model and train it to use that knowledge. We present LM-CORE -- a general framework to achieve this-- that allows \textit{decoupling} of the language model training from the external knowledge source and allows the latter to be updated without affecting the already trained model. Experimental results show that LM-CORE, having access to external knowledge, achieves significant and robust outperformance over state-of-the-art knowledge-enhanced language models on knowledge probing tasks; can effectively handle knowledge updates; and performs well on two downstream tasks. We also present a thorough error analysis highlighting the successes and failures of LM-CORE. Our code and model checkpoints are publicly available\footnote{\url{https://github.com/sumit-research/lmcore}}.

\end{abstract}

\section{Introduction}
\label{sec:intro}

Large pre-trained language models (PLMs)~\citep{elmo, devlin-etal-2019-bert, gpt-3} have achieved state-of-the-art performance on a variety of NLP tasks. Much of this success can be attributed to the significant semantic and syntactic information captured in the contextual representations learned by PLMs. In addition to applications requiring linguistic knowledge, PLMs have also been useful for a variety of tasks involving factual knowledge and it has been shown that models such as BERT~\citep{devlin-etal-2019-bert} and T5~\citep{2020t5} store significant world knowledge in their parameters~\citep{lmaskb}. 

PLMs are typically fed a large amount of unstructured text which leads to the linguistic nuiances and world knowledge being captured in the model parameters. This \textit{implicit storage} of the knowledge in the form of the parameter weights not only leads to poor interpretability while analyzing model predictions but also poses constraints on the amount of knowledge that can be stored. It is not practical to pack all the ever-evolving world knowledge in the language model parameters due to the great financial and environmental costs incurred by training of the PLMs.  Further, since the PLMs acquire knowledge from the text corpora they are trained on, they tend to become sensitive to the contextual and linguistic variations~\cite{whatLMKnow}. Moreover, PLMs do not contain explicit grounding to real world entities, and hence, often find it difficult to \textit{recall} factual knowledge~\citep{logan-etal-2019-baracks}. For example, the model may not be able to \textit{recall} correct information and successfully complete the sentence, ``\textit{The birthplace of Barack Obama is \rule{0.5cm}{0.15mm}}'',  if the LM has seen this fact in a different context  during training (e.g., ``\textit{Barack Obama was born in Honolulu, Hawaii.}'').

 \begin{figure*}[th]
\centering
    \includegraphics[width=0.8\textwidth]{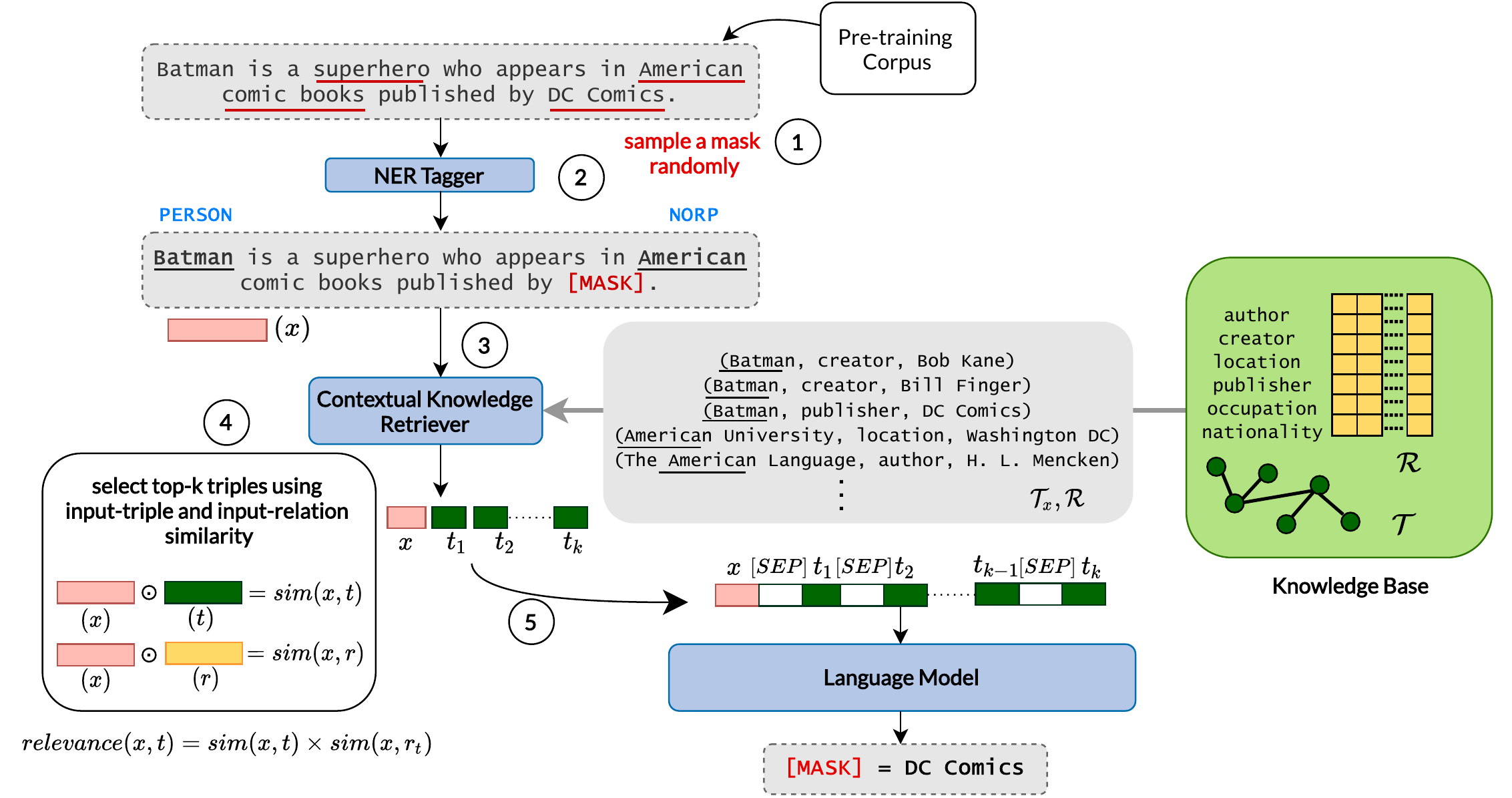}
\caption{\textbf{Language Model Pre-Training with Contextually Relevant External Knowledge:} \circled{1} Using a sentence sampled from the pre-training corpus, an input $(x)$ is created by selecting an entity mention at random from the potential mask candidates (underlined in red). \circled{2} An NER tagger is then applied to the masked input sequence $(x)$ to identify named entities (underlined in black). \circled{3} For the identified entities, the Knowledge Retrieval module fetches the set $\mathcal{T}_x$ of all the triples from the Knowledge Base and then \circled{4} scores all the retrieved triples using input-triple and input-relation similarity (details in Section~\ref{sec:retrieve}). \circled{5} The top-$k$ triples are fed to the Language Model encoder along with the input sequence $(x)$ and the model is trained to predict the masked token. \vspace{-5mm}}
\label{fig:approach}
\end{figure*}

 Large scale structured knowledge bases (KBs) such as YAGO~\citep{10.1145/1242572.1242667} and Wikidata~\citep{wikidata} offer a rich resource of high quality structured knowledge that can provide the PLMs with explicit grounding to real world entities. Consequently, efforts have been made to integrate factual knowledge into PLMs and create entity-enhanced language models~\citep{Peters2019KnowledgeEC, zhang-etal-2019-ernie, colake, kbert, k-adapter, 10.1162/tacl_a_00360}. However, these works either update the PLM parameters or modify the architecture to facilitate the storage of factual knowledge in the model layers and parameters, making it  expensive to update knowledge. 

In this work, we step back and ask -- \textit{what if instead of focusing on storing the knowledge in the language model parameters, we provide the model with \textit{contextually relevant external knowledge} and train it to use this knowledge?} This approach offers several potential advantages -- \textit{(i)} we can utilize the already available high-quality large-scale knowledge bases such as YAGO and Wikidata; \textit{(ii)} not all the knowledge needs to be packed in the parameters of the model resulting in lighter, smaller and greener models; and  \textit{(iii)} as new knowledge becomes available, the knowledge base can be updated independently of the language model.

\noindent
\textbf{Our Contributions:} We present LM-CORE, a framework for augmenting language models with contextually relevant external knowledge. The LM-CORE framework is summarized in Figure~\ref{fig:approach} and consists of a contextual knowledge retriever that fetches relevant knowledge from an external KB and passes it to the language model along with the input text. The language model is then trained with a modified entity-oriented masked language modeling objective
(Section~\ref{sec:approach}). Our proposed solution is simple, yet highly effective. Experiments on benchmark knowledge probes show that the proposed approach leads to significant performance improvements over base language models as well as state-of-the-art knowledge enhanced variants of the language models (Section~\ref{sec:experiments}). We find that with access to contextually relevant external knowledge, LM-CORE is less sensitive to the contextual variations in input text. We also show how LM-CORE can handle knowledge updates without any re-training  and compare the performance of LM-CORE on two knowledge-intensive downstream tasks. Finally, we present an in-depth analysis of cases where our proposed approach gives incorrect answers paving the way for further research in this direction (Section~\ref{sec:error_analysis}).

\section{Related Work}
\label{relatedWork}

\noindent
\textbf{Augmenting Additional Knowledge in PLMs:}
Previous works on augmenting PLMs with additional knowledge can be grouped into two categories. One line of work adopts a retrieve and read framework where the model is trained to retrieve relevant information followed by a reading comprehension step to perform the downstream task~\citep{lee-etal-2019-latent,guu2020realm, agarwal2021knowledge}. 
While our proposal has similarities with this line of work in terms of \textit{retrieving} the contextual knowledge, there are two major differences. First, most of these works consider external knowledge in the form of unstructured text (such as Wikipedia documents). However, extracting factual knowledge from unstructured text is hard and error-prone due to ambiguities in natural language and infrequent mentions of entities of interest~\cite{Peters2019KnowledgeEC}.
This issue can be alleviated by using a structured knowledge base where the knowledge is represented (mostly) unambiguously -- each fact is a triple in the knowledge base. Further, these approaches employ \textit{explicit supervision during pre-training} to train the model to fetch relevant passages from the text. This results in systems that are more complex and resource-hungry than the base PLMs used and also make it difficult to reuse or adapt the models to different sources of knowledge.

The second body of work has focused on injecting the factual knowledge directly into the model parameters by feeding more data to the model during pre-training~\cite{poerner-etal-2020-e, knowledgePackLM}. A promising direction explored recently is utilizing structured knowledge bases to augment Transformer-based LMs. ERNIE~\citep{zhang-etal-2019-ernie} and KnowBERT~\citep{Peters2019KnowledgeEC} are notable efforts in this direction where the entity information from knowledge bases is explicitly linked with the input text during pre-training yielding entity-enhanced variants of BERT models with entity representations integrated within the Transformer layers. An alternative way of training entity-aware language models is illustrated by frameworks such as CoLAKE~\citep{colake}  and KEPLER~\citep{10.1162/tacl_a_00360}  that jointly learn the language and knowledge representations thereby producing language models augmented with factual knowledge and knowledge embeddings enhanced with textual context. However, these approaches, by design, will lead to larger and larger models to store the ever-growing abundant knowledge. Further, due to the strong coupling between the knowledge and language signals, updating or adding knowledge requires re-training of the model.

\noindent
\textbf{Examining the knowledge contained in PLMs:}
\citet{lmaskb} posit that while training over large amounts of input text, PLMs may also be storing (implicit) relational knowledge in their parameters and proposed the Language Model Analysis (LAMA) framework to measure the relational knowledge stored in a PLM.

\citet{whatLMKnow} argue that due to the sensitivity of the PLMs on the input context, such manually created prompts are sub-optimal and might fail to retrieve facts that the PLM \textit{does know}, thus providing only a lower bound estimate of the knowledge contained in it. Subsequent work~\citep{shin-etal-2020-autoprompt, zhong-etal-2021-factual} has attempted to generate better prompts in order to tighten this estimate. \citet{poerner-etal-2020-e} introduced LAMA-UHN (UnHelpfulNames), a much harder subset of LAMA where the input probes provide little or no helpful contextual signals from other tokens in the probe, thus measuring the innate ability of the PLM to recall information.

\section{LM-CORE: Knowledge Retrieval and Training Framework}
\label{sec:approach}
\textbf{Task setting and Overview:} Consider a language model $\mathcal{L}$ (such as BERT and RoBERTa) and a knowledge base $\mathcal{K} = \{t_{hrt} = <h, r, t> | h,t \in \mathcal{E} ; r \in \mathcal{R}\}$. Here, we consider the knowledge base $\mathcal{K}$ as a set of triples such that each triple $t_{hrt}$ represents the relationship $r$ between entities $h$ and $t$. $\mathcal{E}$ is the set of all the entities, and $\mathcal{R}$ is the set of all the relationship types present in the knowledge base. Given a text input $x$, the proposed LM-CORE framework retrieves a set of triples $\mathcal{T}_x \in \mathcal{K}$ such that the triples in $\mathcal{T}_x$ are contextually relevant to $x$. The language model is then presented with the original input $x$ and the contextually relevant knowledge in the form of $\mathcal{T}_x$ and is trained to make predictions using this additional knowledge. 

We posit that the model essentially needs to learn relevant semantic associations between natural language input text and various relation types present in the knowledge base. Identifying the correct relation types will help the model leverage the corresponding relevant facts in order to make an accurate prediction. This is accomplished via a modified Masked Language Modeling (MLM)~\cite{devlin-etal-2019-bert}  pre-training objective. Figure~\ref{fig:approach} summarizes the complete workflow of our proposed LM-CORE framework and we describe the three main components in detail in the following sub-sections. 

\subsection{Entity span masking}
\label{sec:masking}

\begin{figure}[tbp]
\centering
    \includegraphics[width=0.9\linewidth]{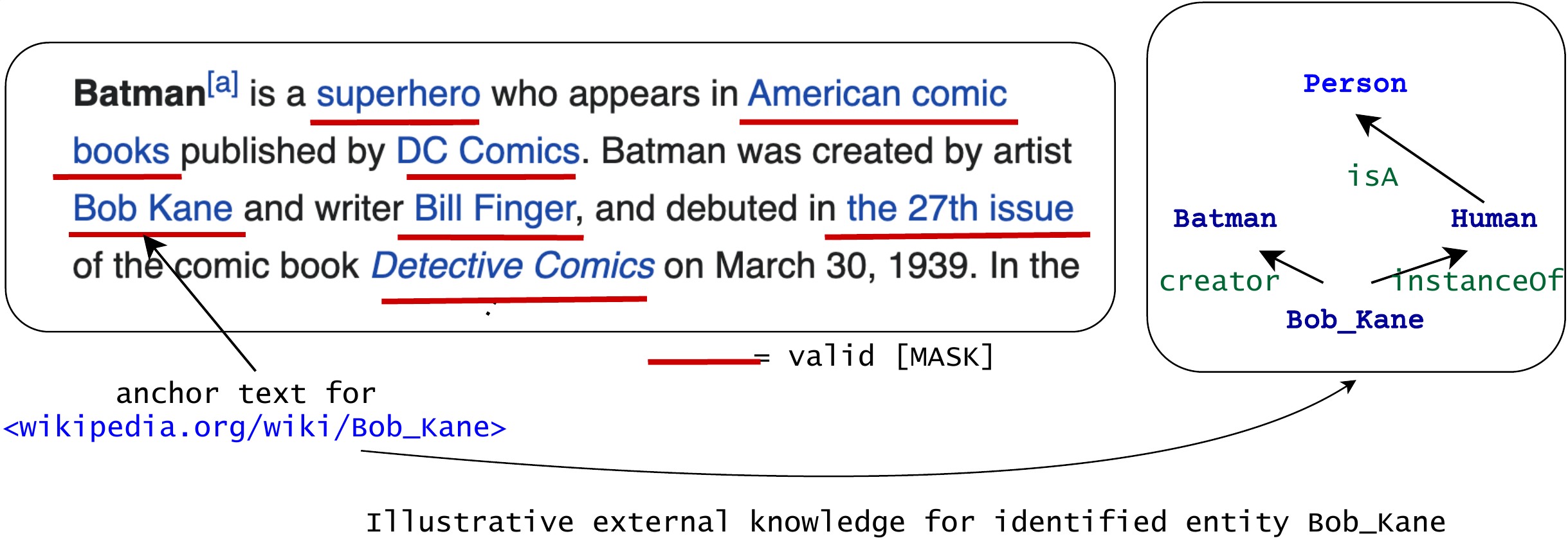}
    \caption{We create our pre-training corpus from Wikipedia by masking entity spans detected using anchor text of hyperlinks.
    }
\label{fig:mask}
\vspace{-1em}
\end{figure}

Masked Language Modelling is a popular task used for training PLMs where the objective is to predict the masked token in the input sequence. In order to improve model's grounding to real world entities, previous works have adopted different strategies for explicitly masking entity information in the input text by using entity representations obtained by knowledge base embeddings~\cite{zhang-etal-2019-ernie}, using named entity recognizers (NER) and regular expressions~\cite{guu2020realm}, and verbalizing knowledge base triples~\cite{agarwal2021knowledge}. These approaches often result in noisy masks due to the limitations of underlying rules, and NER and entity linking systems. To overcome these limitations, we propose a novel way of creating high-quality and accurate entity masks by using Wikipedia as the base corpus for training. Note that in order to create entity masks, we need to identify corresponding entity mentions in the input text for which we utilize the human-annotated links in Wikipedia. The official style guidelines of Wikipedia require the editors to link mentions of topics to their corresponding Wikipedia pages. In Figure~\ref{fig:mask}, the left textbox shows a screenshot of the Wikipedia article about \textsf{Batman} where various other related topics, or concepts, are linked to their corresponding Wikipedia pages (underlined in red in the figure, and displayed as blue anchor-text in Wikipedia). This information provides us with high-quality human annotation of entity mentions in the input text.

As illustrated in Figure~\ref{fig:mask}, the underlined tokens (such as \textsf{DC Comics, Bob Kane, Bill Finger}) constitute the set of entity tokens that could be masked. For each such mask, we can also obtain the corresponding contextual knowledge from the external knowledge base (illustrated for \textsf{Bob Kane} in the right text box). By masking only the entity tokens (instead of randomly sampled words)  and providing contextually relevant knowledge to the model retrieved from the knowledge base (as described in next subsection), we expect the model to learn to predict the masked entity tokens by utilizing the external knowledge.

\subsection{Contextual Knowledge Retrieval}
\label{sec:retrieve}
After preparing the masked input for training, the second component in our framework fetches contextually relevant knowledge to feed to the language model.%
Consider the sentence, ``\textit{Warren Buffet is the chairman of [MASK]}", where the masked token is \textit{Berkshire Hathaway}. In the typical MLM setting, the model only has access to the linguistic and contextual clues present in the input text to predict the masked token. However, if contextually relevant information is available as additional input, the model can use it to output the correct token. 

We consider the problem of finding contextually relevant facts 
given the input query text as an information retrieval (IR) problem and adopt a retrieve and re-rank approach that has empirically been found to perform well in a variety of tasks~\citep{chen-etal-2017-reading, wang2017r3, das2018multistep, yang-etal-2019-end-end}. Recall the example input discussed above -- ``\textit{Warren Buffett is the chairman of [MASK]}". Intuitively, in this input text, there are two important signals that the retriever needs to utilize -- \textit{entity} and \textit{relation} information. The entity mention \textit{Warren Buffett} indicates that we need to fetch facts related to Warren Buffet from the knowledge base. Typically, there are numerous facts related to a given entity in the knowledge base, especially for popular entities such as \textsf{Warren Buffett}. Thus, the retriever also needs to utilize the presence of the word chairman to retrieve facts (KB triples) representing the management or executive relation.

Given an input text, our retriever pipeline performs Named Entity Recognition (NER) to identify named entity mentions in the input text. We use the NER model from FLAIR~\citep{akbik2019flair} to identify named entity mentions and then select KB entities having maximum overlap with the mention-span of the identified entities. For instance, if the input query is ``\textit{Buffett was born in [MASK]}", all of the entities containing Buffett \textsf{- Warren\_Buffett, Howard\_Warren\_Buffett, Howard\_Graham\_Buffett, Volcano\_(Jimmy\_ Buffett\_song)} etc. are selected, but if the query is ``\textit{Warren Buffett was born in [MASK]}", only the first two entities will be chosen). Once these entities are selected, all the facts from the KB involving these entities are retrieved (denoted by $\mathcal{T}_x$ in Figure~\ref{fig:approach}).

After retrieving the facts involving the entities mentioned in the input, we next need to rank these triples based on their relevance to the input. In order to measure the contextual relevance of a given triple $t$ to the input $x$, we compute the following two scores. 
\\ \noindent
\textbf{Query-Triple similarity:} We obtain representations of the input text $x$ as well as the triple $t$ and compute the inner product of the representations to obtain the similarity score as follows.
\begin{equation}
    \begin{small}        
        sim(x, t) = \textsf{Emb}(x)^{T}\textsf{Emb}(t),\;t \in \mathcal{T}_x
    \end{small}
\end{equation}
Here,  \textsf{Emb$(\cdot)$} is obtained using the Sentence Transformer~\citep{reimers-2019-sentence-bert}. While it is straightforward to obtain representations of input $x$, sentence transformer can not be applied directly to KB triples. Application of KB embeddings such as TransE~\cite{transe} is also not feasible as then the representations of the input text and triples will be in different embedding spaces. To overcome this, we adopt a simple approach of verbalizing the knowledge base triples by concatenating the head entity, relationship and the tail entity, and obtain the representation of the verbalized triple from the sentence transformer. For example, the triple \textsf{(Warren\_Buffett, hasOccupation, Investor)} is verbalized as \textit{Warren Buffett has occupation Investor} and is fed as input to the sentence transformer.
\\ \noindent 
\textbf{Relation-based scoring:}  A triple is highly relevant for the input text if the triple represents the same relationship that is being talked about in the text. To capture this intuition, we embed all the relation types in the KB in the same embedding space as triples using the sentence transformer and compute the similarity between the input text and the relation type of the triple as follows.
\begin{equation}
    \begin{small}
    sim(x, r) = \textsf{Emb}(x)^{T}\textsf{Emb}(r),\;r \in \mathcal{R}
    \end{small}
\end{equation}
where $\mathcal{R}$ is the set of all relations in the KB. 
The final relevance score for the triple $t$, $relevance(x, t)$ is obtained by taking a product of the above two scores.

Based on this final score, we select the top-$k$ triples that constitute the contextual knowledge to be fed as input along with $x$ to the LM. We use $k$ = 8 in this work (See Appendix~\ref{app:k-ablation} for effect of varying $k$). Some illustrative examples of the final retrieved knowledge base triples are presented in Appendix~\ref{app:triple-examples}.

\subsection{Language Model Pre-training with Contextual Knowledge}
\label{pre-train}
With the masked training corpus and the module to fetch contextually relevant knowledge, we now train the model to utilize the additional contextual knowledge to predict the masked token. From the masked corpus, we select a sentence and a valid entity span is chosen at random out of all the potential spans in the sentence. We mask this span to create the input text $x$.  We filter out sentences starting with pronouns such as \textit{he, she, her}, and \textit{they} as we observed that most of such sentences do not contain other useful signals to unambiguously predict the masked words. For instance, if the input example is - ``\textit{He developed an interest in investing in his youth, eventually entering the Wharton School of the University of Pennsylvania}" and \textit{Wharton School of the University of Pennsylvania} is masked, the remaining words in the sentence are not providing any informative signals to the model to predict the masked tokens. Given the input sentence thus selected, the contextual knowledge retriever fetches the relevant triples from the knowledge base. The representations of the input sentence and the retrieved triples are then concatenated and fed to the model and the model is trained to minimize the following MLM loss. 
\begin{equation}
    \begin{small}
        L_{MLM} = \frac{1}{M}\sum_{m\:=\:1}^{M} \log p(x_{{ind}_m} \mid x, t_1, t_2, ..., t_k) 
    \end{small}
\end{equation}
where $M$ is the total number of \textsf{[MASK]} tokens in $x$ and $ind_m$ is the index of the $m^{th}$ masked token.

With the additional contextual information available to the model, we expect the model to learn the associations between linguistic cues in the input text and relevant relationship information in the triples. For example, we expect the model to associate different ways in which someone's date of birth could be mentioned in natural language (such as \textit{X was born on}, \textit{the birthday of X is},  and numerous other linguistic variations) to the  KB relation \textsf{birthDate} and utilize the information from the corresponding triple. Note that since the types of relations in the knowledge base are relatively small in number, and do not change often, we expect the model to generalize well and be more robust to linguistic variations. 

\section{Experiments and Discussions}
\label{sec:experiments}
\noindent
\textbf{Data Sources and Pre-processing:}
\begin{table*}[t]
\centering
\caption{Mean precision at one (P@1) of various models on LAMA probe. We group all the models based on the base language model used (BERT or RoBERTa). For LM-CORE, $(\cdot,\cdot)$ indicates the variant --  (b, r corresponds to BERT and RoBERTa, respectively, and y, w indicate YAGO and Wikidata5M, respectively). Best results in each column are highlighted in bold and the second best performance is underlined}
\vspace{-0.68em}
\label{table:LAMA}
\resizebox{\textwidth}{!}{%
\begin{tabular}{@{}lcccccllllll@{}}
\toprule
 & \multirow{2}{*}{\textbf{Complete}} & \multicolumn{4}{c}{\textbf{Google-RE}} & \multicolumn{4}{c}{\textbf{T-REx}} & \multicolumn{1}{c}{\multirow{2}{*}{\textbf{SQuAD}}} & \multicolumn{1}{c}{\multirow{2}{*}{\textbf{\begin{tabular}[c]{@{}c@{}}Concept\\ Net\end{tabular}}}} \\
 \cmidrule(r){3-6} \cmidrule(r){7-10}
 &  & \textbf{DoB} & \textbf{PoB} & \textbf{PoD} & \textbf{All} & \multicolumn{1}{c}{\textbf{1-1}} & \multicolumn{1}{c}{\textbf{N-1}} & \multicolumn{1}{c}{\textbf{N-M}} & \multicolumn{1}{c}{\textbf{All}} & \multicolumn{1}{c}{} & \multicolumn{1}{c}{} \\
 \midrule
 \midrule
\multicolumn{12}{@{}l}{\textit{BERT-based models}} \\
\midrule
\textbf{BERT-base} & 24.73 & \underline{1.59} & 15.46 & 10.33 & 9.12 & 67.94 & 32.67 & 23.54 & 30.83 & 14.29 & 15.88 \\
\textbf{BERT-large} & 25.44 & \underline{1.59} & 15.53 & 12.16 & 9.76 & 74.23 & 31.30 & 25.30 & 31.05 & \textbf{17.61} & \textbf{18.72} \\
\textbf{ERNIE} & 22.16 & 1.42 & 13.48 & 4.97 & 6.62 & 61.51 & 28.57 & 21.93 & 27.58 & 13.62 & 14.83 \\
\textbf{LM-CORE(b,y)} & \underline{39.64} & \textbf{64.44} & \textbf{52.71} & \textbf{50.98} & \textbf{56.04} & \underline{74.37} & \underline{51.18} & \underline{34.57} & \underline{45.83} & 15.61 & 14.78 \\
\textbf{LM-CORE(b,w)} & \textbf{42.83} & 0.66 & \underline{37.62} & \underline{31.11} & \underline{23.13} & \textbf{81.79} & \textbf{59.86} & \textbf{45.48} & \textbf{55.32} & \underline{17.28} & \underline{16.15}  \\
 \midrule
 \midrule
\multicolumn{12}{@{}l}{\textit{RoBERTa-based models}} \\
\midrule

\textbf{RoBERTa} & 20.46 & \underline{1.85} & 12.98 & 1.23 & 5.35 & 57.49 & 23.14 & 21.59 & 24.21 & 12.94 & \underline{18.47} \\
\textbf{RoBERTa-large} & 24.24 & 1.41 & 12.48 & 0.46 & 4.78 & 70.24 & 29.08 & 23.28 & 28.82 & \textbf{18.88} & \textbf{22.09} \\
\textbf{KEPLER} & 19.36 & 1.47 & 11.73 & 3.08 & 5.43 & 52.32 & 21.58 & 21.41 & 23.01 & 9.10 & 17.25 \\
\textbf{CoLAKE} & 23.38 & 1.79 & 15.72 & 10.79 & 9.43 & 64.08 & 29.40 & 23.54 & 28.80 & 8.39 & 17.17 \\
\textbf{LM-CORE(r,y)} & \underline{34.60} & \textbf{46.33} & \textbf{43.47} & \underline{26.35} & \textbf{38.71} & \underline{68.21} & \underline{45.30} & \underline{30.40} & \underline{40.60} & 13.29 & 17.53 \\
\textbf{LM-CORE(r,w)} & \textbf{41.96} & 0.38 & \underline{33.11} & \textbf{28.20} & \underline{20.56} & \textbf{70.21} & \textbf{60.30} & \textbf{43.18} & \textbf{54.11} & \underline{15.73} & 18.38
\\ \bottomrule
\end{tabular}%
}
\vspace{-1em}
\end{table*} 
We create our pre-training corpus using the December 20, 2018 snapshot of English Wikipedia that contains about 5.5M documents. Processing all the articles following the masking strategy described in Section~\ref{sec:masking} resulted in a total of $\sim$46.3M sentences with valid masks, from which we randomly sample sentences to create input examples. 

In order to illustrate the general nature of LM-CORE, we used two different PLMs as our LM encoders -- BERT-base (uncased) model and RoBERTa-base (cased) model. We use YAGO and Wikidata as two different knowledge bases giving us four variants of LM-CORE ($\{bert, roberta\} \times \{yago, wikidata\}$). We use the English Wikipedia version of YAGO 4~\citep{10.1145/1242572.1242667} and preprocess it to obtain our retrieval corpus consisting of roughly 17M triples spanning over 4.9M entities and 131 unique relations. For Wikidata, we used the Wikidata5M version~\cite{10.1162/tacl_a_00360} that consists of roughly 21M triples covering 821 unique relations and 4.8M entities. Further details regarding retrieval corpus generation and processing can be found in the Appendix (Section \ref{app:rt-corpus}). For computing triple representations for retrieval (Section~\ref{sec:retrieve}), we concatenate the subject (head), relation, and object (tail) of triples and embed them using the Sentence Transformers~\citep{reimers-2019-sentence-bert} and obtain the 768-dimensional embeddings (same as LM encoder dimensions).

\subsection{Does External Knowledge Help PLMs in Knowledge Intensive Tasks?}
We now present an analysis of how much, and if, having access to external knowledge can help PLMs in knowledge-intensive tasks. A popular way of assessing a model's ability to perform at such tasks is by using benchmark knowledge probes. We use the LAMA probe~\citep{lmaskb}

Table~\ref{table:LAMA} reports the performance of various PLMs on the LAMA probe as measured by Precision at 1 (P@1). The numbers in the Table are grouped based on the base language model used by different models. We use ERNIE~\cite{zhang-etal-2019-ernie} (based on BERT and Wikidata), and KEPLER~\cite{10.1162/tacl_a_00360} and CoLAKE~\cite{colake} (based on RoBERTa) as the representative knowledge enhanced language models. Both KEPLER and CoLAKE have used Wikidata5M as the knowledge base. We used author provided code and checkpoints for obtaining the reported numbers. For LM-CORE, we use four variants with different knowledge base and language encoder combinations as described above.

We observe that our approach of providing external knowledge to the PLMs leads to substantially improved performance over the base language models and their SoTA knowledge enhanced variants. LM-CORE(b,w) achieves P@1 of 42.83\% compared to 25.44\% for BERT-large. Likewise, LM-CORE(r,w) achieves a P@1 of 41.96\% significantly outperforming RoBERTa-large (24.24\%). We also report the numbers on the four different subsets of LAMA revealing interesting insights. For all the models considered, we note that the performance on T-REx subset is higher than the Google-RE subset. We attribute this to the nature of knowledge required for probes in the four subsets. Note especially the column for Date of Birth (DoB) in the Table. All the models, except for LM-CORE(b,y) and LM-CORE(r,y) perform extremely poorly. This is because the Wikidata5M KB does not have \textit{date} entity type and hence, the poor performance of models using Wikidata.  We also note that on the SQuAD and Concept Net subsets, the knowledge enhanced models do not offer significant improvements over the base language models. While Google-RE and T-REx focus more on factual world knowledge (present in abundance in YAGO and Wikidata), SQuAD and ConceptNet concentrate more on commonsense knowledge (limited in YAGO and Wikidata). This is a major focus of our continuing work on enhancing the external knowledge with commonsense knowledge bases such as ConceptNet~\citep{conceptnet} Atomic~\citep{atomic}.

\subsection{Sensitivity to Contextual Signals in Input}
\label{sec:lama-uhn}
\begin{table}[t]
\centering
\caption{P@1 for different models on LAMA-UHN.}
\vspace{-0.68em}
\label{table_LAMA-uhn}
\begin{small}
\begin{tabular}{@{}l@{\hskip 3em}lll@{}}
\toprule
 &  \textbf{LAMA} & \textbf{\begin{tabular}[c]{@{}l@{}}LAMA\\ UHN\end{tabular}} & \textbf{\begin{tabular}[c]{@{}l@{}}Percentage\\Change\end{tabular}}  \\
 \midrule
 \midrule
\multicolumn{4}{@{}l}{\textit{BERT-based models}} \\
\midrule
\textbf{BERT-base} & 24.73  & 18.72 & -24.31   \\
\textbf{BERT-large} & 25.44 & 19.92  & -21.67\\
\textbf{ERNIE} & 22.16 &  15.81  & -28.66\\
\textbf{LM-CORE(b,y)} & 39.64 &  \underline{41.33}  & +4.26\\
\textbf{LM-CORE(b,w)} & 42.83 &  \textbf{45.50}  & +6.23\\
\midrule 
 \midrule
\multicolumn{4}{@{}l}{\textit{RoBERTa-based models}} \\
\midrule
\textbf{RoBERTa-base} & 20.46  & 13.66  & -33.24  \\
\textbf{RoBERTa-large} & 24.24 & 17.99  &  -25.78\\
\textbf{KEPLER} & 19.36 & 12.46  & -35.64  \\
\textbf{CoLAKE} & 23.38 & 17.16 &  -13.74\\
\textbf{LM-CORE(r,y)} & 34.60  & \underline{34.25}  & -1.01 \\
\textbf{LM-CORE(r,w)} & 41.96 & \textbf{44.75}  & +6.65
\\ \bottomrule
\end{tabular}
\end{small}
\vspace{-1.em}
\end{table}
PLMs are often sensitive to the linguistic variations in the input and are overly reliant on the surface form of entity names for making its predictions~\citet{poerner-etal-2020-e}. For example, BERT can predict that a person with an Italian-sounding name was born in Italy even if this is factually incorrect. In order to evaluate the sensitivity and robustness of different models, we report the P@1 numbers for the LAMA-UHN (UnHelpfulNames) probing benchmark (Table~\ref{table_LAMA-uhn}) -- a much harder subset of LAMA where input probes with helpful entity names are removed and the PLM has little or no helpful contextual signals from other tokens in the probe. 
We observe that the LM-CORE variants significantly outperform the base language models and their knowledge enhanced variants. Further, note that while all the baseline models suffer a significant fall in performance (expected due to the hardness of LAMA-UHN), the drop in performance of LM-CORE variants is much less. This indicates that having access to relevant external knowledge helps reduce the dependence on linguistic signals and results in the \textit{robust} outperformance of LM-CORE variants.

\begin{figure}
    \centering
    \includegraphics[width=0.9\columnwidth]{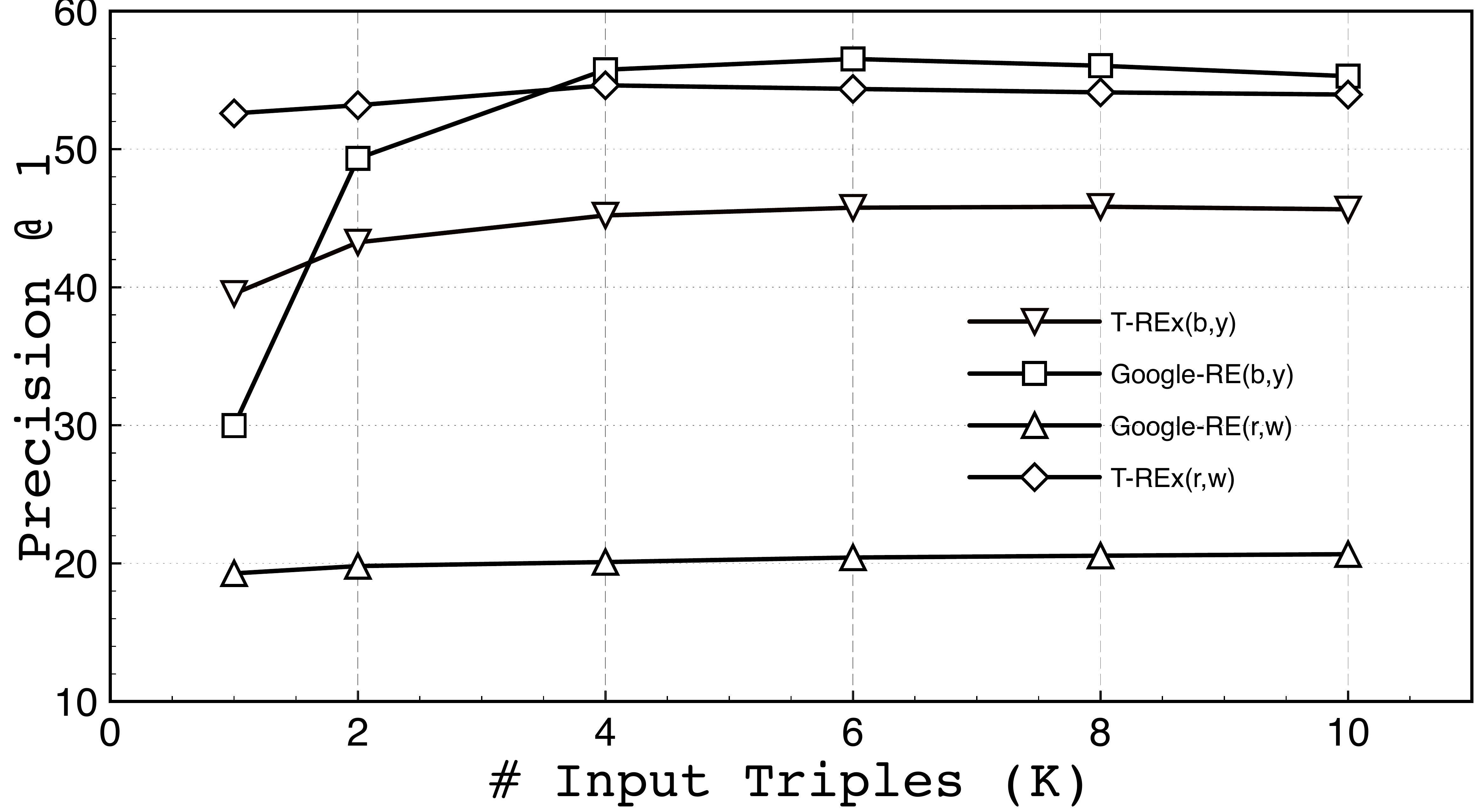}
    \caption{Effect of $k$ on performance of different models.}
    \label{fig:k-ablation}
\end{figure}

\subsection{Effect of Varying Number of Input Triples to LM-CORE}
\label{app:k-ablation}

We analyze the effect of varying the number of candidates ($k$) during retrieval in Figure \ref{fig:k-ablation}. We discuss with respect to Google-RE and T-REx subsets as our factual knowledge triples are most relevant for answering queries in these subsets (in comparison to commonsense queries in ConceptNet).

We plot the Precision@1 (P@1) against increasing $k$ values from 1 to 10 for LM-CORE(b,y) and LM-CORE(r,w) variants. We do not observe any consistent optimal $k$ value across variants and data subsets. To add, there is no significant difference between P@1 values as $k$ varies from 4 to 10. Hence, in order to maximize our recall while keeping the computational expense in mind, we select $k=8$ for our experiments.

\subsection{Role of LM-CORE Pre-training and Retrieved Knowledge}

We now study the role LM-CORE pre-training plays in helping the model access and utilize the retrieved knowledge and ensure that the model does not just rely on the knowledge stored in its parameters. We also study the effect of augmenting the base LMs with knowledge retrieved by LM-CORE. In addition to providing an insight into the quality of the knowledge retrieved by LM-CORE, this will also help us better understand the ability of LM-CORE to utilize the retrieved knowledge. 

We consider the following four variants on the LAMA probe (Table \ref{table_ablation_use_kb}):
\begin{enumerate}
\itemsep0em 
    \item RoBERTa-base 
    \item RoBERTa-base + triples retrieved by LM-CORE’s Contextual Knowledge Retriever
    \item LM-CORE(r,w) + random triples
    \item LM-CORE(r,w)
\end{enumerate}

We observe that LM-CORE(r,w)’s performance (41.69 P@1) significantly exceeds RoBERTa-base’s performance using the same triples in input (30.06 P@1) , demonstrating that our training procedure equips the model with the capability of identifying and using relevant external knowledge effectively. There is a large drop in performance (from 41.69 P@1 to 19.51 P@1) when LM-CORE(r,w) is provided with random triples in input. This shows that the model exclusively accesses external knowledge to answer queries correctly. While the performance drops, it is important to note that the P@1 is similar to RoBERTa-base (20.46 P@1), highlighting that our training procedure does not lead to \textit{catastrophic forgetting} and the model is able to rely on the knowledge stored in its parameters when semantically relevant triples are not provided in the input.
Finally, although RoBERTa-base when augmented with contextually relevant triples does not perform competitively with LM-CORE, it demonstrates considerable improvement over the base RoBERTa model. This shows that high-quality relevant external knowledge has the potential to improve factual prediction, further reinforcing our motivation to train models to efficiently retrieve and use this knowledge.

\begin{table}[t]
\centering
\caption{Ablation study analyzing the effectiveness of LM-CORE pre-training and contextually retrieved knowledge (Precision@1 values).}
\label{table_ablation_use_kb}
\begin{small}
\resizebox{\columnwidth}{!}{
\begin{tabular}{@{}llllll@{}}
\toprule
 &  \textbf{All} & \begin{tabular}[c]{@{}l@{}}\textbf{Google}\\ \textbf{RE}\end{tabular} & \textbf{T-REx} & \textbf{SQuAD} & \begin{tabular}[c]{@{}l@{}}\textbf{Concept}\\ \textbf{Net}\end{tabular} \\ \midrule
\begin{tabular}[c]{@{}l@{}}\textbf{RoBERTa}\\ \textbf{(base)}\end{tabular}  & 20.46 & 5.35 & 24.21 & 12.94 & 18.47  \\  
\begin{tabular}[c]{@{}l@{}}\textbf{RoBERTa-base}\\ \textbf{+LM-CORE triples}\end{tabular}  & 30.06 & 9.79 & 38.71 & 7.69 & 16.26  \\
\midrule
\begin{tabular}[c]{@{}l@{}}\textbf{LM-CORE(r,w)} +\\ \textbf{random triples}\end{tabular}  & 19.51 & 9.05 & 22.74 & 13.99 & 16.92  \\
\textbf{LM-CORE(r,w)}  & 41.69 & 20.56 & 54.11 & 15.73 & 18.38  \\ \bottomrule
\end{tabular}
}
\end{small}
\end{table}

\subsection{Downstream Tasks}
We consider two downstream tasks to study the effectiveness of LM-CORE for different NLP applications. We take Zero-Shot Relation Extraction (ZSRE)~\cite{zsre} and open-domain question answering over Web Questions (WQ)~\cite{webqa} dataset as the representative knowledge-intensive tasks. Tables~\ref{tab:zsre} and ~\ref{tab:webq} report the performance of LM-CORE and various other baselines for the two tasks, respectively. We use the LM-CORE(b,w) variant for these experiments as most baselines use BERT as the LM and Wikipedia as the knowledge base. For the ZSRE task, we use the data splits and evaluation systems provided as part of the KILT benchmark~\citep{kilt}. We find that for the ZSRE task, LM-CORE achieves a significantly higher F-1 score (\textbf{74.80}) compared to the second-best RAG model (49.95). Also, note that the online evaluator for the task considers exact string match (including casing, punctuations, etc.) for computing accuracy numbers but not for computing other metrics. Hence, the reported accuracy number for LM-CORE represent a lower bound as we don't have access to the same pre-processing pipeline to process its output. For the WQ dataset, we find that LM-CORE outperforms BERT with BM25 and neural retrievers, and the DrQA system. We observe that LM-CORE is outperformed by ORQA, designed explicitly for this task, and RAG (a retrieval augmented generative model). However, do note that all the models except LM-CORE have access to much larger knowledge source (complete Wikipedia corpus. $\approx$ 2B words), whereas LM-CORE only has access to the KB triples (21M triples, $\approx$ 140M words). As we show in the following subsection, with access to additional external knowledge, the performance of LM-CORE can improve significantly.  

\begin{table}[h]
\centering
\caption{F1 and Accuracy on Zero Shot RE. $^*$ The accuracy for LM-CORE is the lower-bound number as the online evaluator considers exact string match to compute accuracy.}
\vspace{-0.7em}
\label{tab:zsre}
\begin{small}
\resizebox{\columnwidth}{!}{
\begin{tabular}{@{}l@{\hskip 3em}lll@{}}
\toprule
 &  \textbf{\#params} & \textbf{F1} & \textbf{Accuracy}  \\
 \midrule
\textbf{BERT+DPR~\cite{dpr}} & 330M  & 37.28 & 6.93  \\
\textbf{T5 (base)} & 220M &  13.52  & 9.02\\
\textbf{BART (large)} & 406M & 12.21  & 9.14\\
\textbf{BART+DPR} & 626M &  34.47  & 30.43\\
\textbf{RAG~\cite{rag}} & 626M &  49.95  & \textbf{44.74}\\
\midrule
\textbf{LM-CORE(b,w)} & 110M  &  \textbf{74.80}  & 14.24$^*$\\
\bottomrule
\end{tabular}
}
\end{small}
\vspace{-1em}
\end{table}

\begin{table}[t]
\centering
\caption{Accuracy on Web Questions. BM25, Neur. Retriever, DRQA and LM-CORE perform static retrieval.}
\vspace{-0.7em}
\label{tab:webq}
\begin{small}
\begin{tabular}{@{}lll@{}}
\toprule
\multicolumn{1}{l}{} & Accuracy \\ \midrule
\textbf{BM25 + BERT} & 17.7 \\
 \textbf{Neur. Retriever + BERT} & 7.3 \\
 \textbf{DrQA~\citep{chen-etal-2017-reading}} & 20.7 \\ 
 \textbf{LM-CORE (b,w)} & 21.9 \\
\midrule
\textbf{ORQA~\citep{orqa}} & 36.4 \\ 
\textbf{RAG~\citep{rag}} & 45.5 \\ 
\bottomrule
\end{tabular}
\end{small}
\vspace{-1.5em}
\end{table}

\subsection{Handling Knowledge Updates} 
\label{sec:static}
Once PLMs have been trained, it is expensive to retrain them with new and updated knowledge. LM-CORE, on the other hand, can easily access updated knowledge as it is \textit{external} to the model. Instead of storing all the KB facts in the PLM, LM-CORE essentially learns semantically relevant associations between the input text and KB relations and can easily \textit{query} for the relevant knowledge by using the learned relationship associations. We illustrate this ability to handle dynamic knowledge by introducing new triples in the KB and verifying if LM-CORE is able to leverage this new information to correct its earlier predictions. We consider the LM-CORE(b,y) variant for this experiment. We randomly sample 100 instances from the LAMA probe where the model failed and manually analyze these instances to identify the cases where the corresponding fact was not present in the YAGO KB. There were a total of of 41 such instances and we manually added the correct facts needed to answer the corresponding questions in YAGO. We then presented the 41 inputs again to the model with the updated KB. This time, the model used this newly added knowledge and was able to correct its prediction \textit{without any re-training} for 36 out of 41 cases (87.8\%). As discussed in the following sub-section (\ref{sec:error_analysis}), a majority of errors made by LM-CORE are due to missing facts in the KB and we expect that most of such errors can be corrected by having access to a larger, more comprehensive Knowledge Base.

\subsection{Discussions}
\label{sec:error_analysis}
We now present some representative examples to illustrate the successes and failures of LM-CORE.
Consider a test probe from the Google-RE subset of LAMA -- \textit{Phil Mogg is a member of} \underline{\hspace{2em}}. Here, the correct output token is \textit{UFO}, the band and BERT model incorrectly predicts \textit{parliament} as the output token. This highlights the sensitivity of PLMs on context; BERT's prediction seems to be derived from its memorization of the frequently encountered phrase \textit{member of parliament} during pre-training. We argue that the contextual knowledge retrieved by LM-CORE which includes the relevant fact $<$\textit{Phil Mogg; member of; UFO (band)}$>$ has helped the model to produce the correct output. We present more such successful examples in the Appendix (Tables~\ref{table:LAMA_analysis_bert} and~\ref{table:LAMA_analysis_roberta}).

Next, we analyzed the cases where the proposed framework produced incorrect output and observed three major reasons for erros -- \textit{(i)} the required knowledge was not present in the knowledge base; \textit{(ii)} the required knowledge was not retrieved despite being present in the knowledge base; and \textit{(iii)} the system made errors after retrieving the relevant knowledge. The first problem cause could be addressed by enhancing the knowledge base as shown in Section~\ref{sec:static}. The other two causes of failure highlight the scope of improvement in our retrieval module as well as pre-training module, where further training could help the model make better use of the retrieved knowledge. Some representative examples of these different cases are presented in the Appendix (Table \ref{table:LAMA_error_analysis}). Finally, we noticed some errors that could be attributed to the characteristics of the LAMA probe. Specifically, there are input probes that refer to entities without providing any additional context for disambiguation. For example, the sentence ``\textit{James Johnson was born in \rule{0.5cm}{0.15mm}}" has no clues to determine whether the prompt is referring to the basketball player, Virginia congressman, or the Governor of Georgia with this name. We also noticed certain probes where there are multiple correct completions and the benchmark considers only one of these as the correct answer. For example, ``\textit{Michelangelo is a \rule{0.5cm}{0.15mm} by profession}" can be correctly completed by \textit{poet, painter} or \textit{architect}, but the evaluation considers only \textit{poet} as the correct answer. We also noticed some input examples with highly unambiguous language. For example, ``\textit{X died in \rule{0.5cm}{0.15mm}}", can refer to either X's place of death or date of death but only the former is accepted as the correct answer. Lastly, there are cases where slight (and correct) variations of the expected answer are evaluated as incorrect by the probe. For example, for the prompt ``\textit{Harashima is \rule{0.5cm}{0.15mm} citizen.}" \textit{Japan} is provided as the correct answer while the prediction made by LM-CORE (\textit{Japanese}) is considered incorrect.

\section{Conclusion}
We presented LM-CORE, a framework to train language models with contextually relevant external knowledge. We show that having access to external knowledge leads to significant and robust outperformance over base language models and their knowledge enhanced versions on knowledge probing and two downstream tasks. We also showed how LM-CORE can handle knowledge updates and presented a thorough error analysis that helped us identify possible directions of future work.

\bibliography{refs}
\bibliographystyle{acl_natbib}

\section*{Appendix}
\label{sec:appendix}
\appendix

\section{LM-CORE Training Details}
\label{app:pt-details}
We used the Hugging Face Transformer\footnote{\url{https://github.com/huggingface/transformers}} models BERT and RoBERTa as base models for LM-CORE pre-training. We use the \texttt{BASE} (12-layer, 768-hidden, 12-heads) size models and initialize from \texttt{bert-base-uncased} and \texttt{roberta-base} parameters, respectively. This is consistent with the initialization in~\citet{Peters2019KnowledgeEC, zhang-etal-2019-ernie, 10.1162/tacl_a_00360} and is a common practice to save up on pre-training time.

We used the Adam~\cite{DBLP:journals/corr/KingmaB14} optimizer and a learning rate of 3e-5 across all settings. We could not perform a lot of hyperparameter tuning owing to the computational requirements of the task. Pre-training was done using 8 Nvidia A100 GPUs with a batch size of 512 using gradient accumulation. The masked LM loss continued to decrease at the end of pre-training, suggesting further improvement in performance can be expected. The pre-trained checkpoints for all four variants of LM-CORE can be found \href{https://drive.google.com/drive/folders/1Y3h6OWUcO-3b6UTmU8PFVl0OTRZ6G78a?usp=sharing}{here}.

\subsection{Pre-Training corpus}
\label{app:pt-corpus}
We use the English Wikipedia (December 20, 2018) snapshot\footnote{\url{https://archive.org/download/enwiki-20181220/enwiki-20181220-pages-articles.xml.bz2}} to create our pre-training corpus and WikiExtractor\footnote{\url{https://github.com/attardi/wikiextractor}} to process the dumps. This Wikipedia version contains about 5.5M documents. We retain the hyperlinks while extracting Wikipedia articles as we use them for creating entity masks (Section \ref{sec:masking}). Following the entity masking strategy described in Section \ref{sec:masking}, we obtain our pre-training corpus which contains $\sim$46.3M sentences in total. 

During pre-training the base LMs, we sample sentences containing valid masks. The pre-training corpus is maintained consistent across all LM-CORE variants.

\subsection{Retrieval corpus}
\label{app:rt-corpus}
We use two popular knowledge bases (KBs) in LM-CORE - YAGO 4~\cite{10.1145/1242572.1242667} and Wikidata5M~\cite{10.1162/tacl_a_00360}. The statistics of the KBs -- number of facts, entities and relations can be found in Table~\ref{table_KB-numbers}. We describe the preprocessing steps followed to obtain the respective final retrieval corpora in the following subsections.

\subsubsection{YAGO}

YAGO 4\footnote{\url{https://yago-knowledge.org/downloads/yago-4}} is in RDFS format. YAGO facts are derived from Wikidata, however, all the entities are arranged in a taxonomy mapped to \href{https://schema.org/}{schema.org}.

We pre-process YAGO to remove triples involving relationships such as \textit{image, logo} and \textit{url} that point to meta-data such as images and other files. We also filter out triples that point to RDF literals or Wikidata URLs. 

\subsubsection{Wikidata5M}
We use the Wikidata5M subset of Wikidata as made available by \citet{10.1162/tacl_a_00360}~\footnote{\url{https://deepgraphlearning.github.io/project/wikidata5m}}. This subset of Wikidata is aligned with Wikipedia such that each entity in Wikidata5M has a corresponding entry in Wikipedia. We used the \texttt{raw} graph as provided in the dataset, the statistics of which are reported in Table~\ref{table_KB-numbers}.

\begin{table}[h]
\centering
\caption{Statistics of the knowledge bases used for retrieval in terms of number of triples, entities and relations}
\label{table_KB-numbers}
\begin{small}
\begin{tabular}{@{}l@{\hskip 2em}lll@{}}
\toprule
 & \textbf{Facts} & \textbf{\begin{tabular}[c]{@{}l@{}}Entities\end{tabular}} & \textbf{\begin{tabular}[c]{@{}l@{}}Relations\end{tabular}} \\
 \midrule \\
\textbf{Yago} & 17,421,942 & 4,927,897 & 131 \\
\textbf{Wikidata5M} & 21,285,880 & 4,797,808 & 821 
\\ \bottomrule
\end{tabular}
\end{small}
\end{table}

\subsection{Example Retrieved Triples}
\label{app:triple-examples}
\begin{table*}[t]
\centering
\resizebox{\textwidth}{!}{%
\begin{tabular}{@{}lll@{}}
\toprule
\textbf{Input} & \textbf{Masked Token} & \textbf{Candidates}  (correct fact in bold) \\ \midrule
\multirow{3}{*}{\begin{tabular}[c]{@{}l@{}}Henri Jules Louis Marie Rendu \\ (24 July 1844 – 16 April 1902) was \\ a French physician born in {[}MASK{]}.\end{tabular}} & \multirow{3}{*}{Paris} & (Henri Jules Louis Marie Rendu; birth date; 1844-07-24) \\
 &  & \textbf{(Henri Jules Louis Marie Rendu; birth place; Paris)} \\
 &  & (Henri Jules Louis Marie Rendu; death date; 1902-04-16) \\
 &  & (Henri Jules Louis Marie Rendu; nationalit;y France) \\
 &  & (Henri Jules Louis Marie Rendu; given name; Henri) \\
 \midrule
\multirow{3}{*}{Weisenborn attended the {[}MASK{]}.} & \multirow{3}{*}{\begin{tabular}[c]{@{}l@{}}University\\of Chicago\end{tabular}} & \textbf{(Gordon Weisenborn; alumni of; University of Chicago)} \\
 &  & (Clara Weisenborn; member of; Republican Party (United States) \\
 &  & (G\"{u}nther Weisenborn; nationality; Germany) \\
 &  & (G\"{u}nther Weisenborn; death place; West Berlin) \\
 &  & (Clara Weisenborn; nationality; United States) \\
 \midrule
\multirow{3}{*}{\begin{tabular}[c]{@{}l@{}}Dehorokkhi (English: Bodyguard)\\ is a Bangladeshi {[}MASK{]} directed\\ by Iftakar Chowdhury.\end{tabular}} & \multirow{3}{*}{action film} & (Dehorokkhi; director; Iftakar Chowdhury) \\
 &  & (Dehorokkhi; in language; Bengali language) \\
 &  & \textbf{(Dehorokkhi; genre; Action film)} \\
 &  & (Bangladeshi Idol; in language; Bengali language) \\
 &  & (British Bangladeshi Who’s Who; in language; English language) \\
\midrule
\multirow{3}{*}{\begin{tabular}[c]{@{}l@{}}Palaemon macrodactylus is \\ a {[}MASK{]} of shrimp of the \\family Palaemonidae.\end{tabular}} & \multirow{3}{*}{species} & (Palaemon macrodactylus; parent taxon Palaemon (genus)) \\
 &  & (Palaemon macrodactylus; parent taxon; Palaemon (genus)) \\
 &  & \textbf{(Palaemon macrodactylus; taxonomic rank; Species)} \\
 &  & (Palaemonidae; taxonomic rank; Family (biology)) \\
&  & (Palaemonidae; parent taxon; Palaemonoidea) \\ 
\bottomrule
\end{tabular}%
}
\caption{Examples of masked input sentences (from Wikipedia) and top-$5$ retrieved candidates during pre-training. }
\label{table:input_cands_ex}
\end{table*}

We provide a closer look into our pre-training approach by showing examples of masked input sentences and the retrieved triple candidates from the knowledge base (Table \ref{table:input_cands_ex}). We observe that the facts retrieved are highly relevant for predicting the masked entities in the input context.

\section{Additional Experiments}

\subsection{How LM-CORE Compares with Other Retrieval Paradigms}

Table~\ref{table:realm} also reports results of REALM~\citep{guu2020realm} -- a retrieval-based language model that retrieves relevant documents from a text corpus during pre-training. We observe that LM-CORE outperforms REALM on the ConceptNet, DoB (Google-RE), and 1-1 (T-REx) subsets, while REALM outperforms the proposed solution in other subsets of the LAMA probe. We specifically highlight an absolute 15 points improvement on the \texttt{date-of-birth} relation despite REALM using explicit date masks while training whereas our training corpus only has entity masks. This indicates that our model can use the the contextual knowledge provided by the retriever module even though it is not explicitly shown such knowledge during training.
\begin{table*}[t]
\centering
\caption{Precision at one (P@1) on LAMA probe. We consider only LM-CORE(b,y) and LM-CORE(b,w) as REALM is trained on BERT. Best results in each column are highlighted in bold and the second best performance is underlined.}
\label{table:realm}
\resizebox{\textwidth}{!}{%
\begin{tabular}{@{}lccccllllll@{}}
\toprule
  & \multicolumn{4}{c}{\textbf{Google-RE}} & \multicolumn{4}{c}{\textbf{T-REx}} & \multicolumn{1}{c}{\multirow{2}{*}{\textbf{SQuAD}}} & \multicolumn{1}{c}{\multirow{2}{*}{\textbf{\begin{tabular}[c]{@{}c@{}}Concept\\ Net\end{tabular}}}} \\
 \cmidrule(r){2-5} \cmidrule(r){6-9}
 &   \textbf{DoB} & \textbf{PoB} & \textbf{PoD} & \textbf{All} & \multicolumn{1}{c}{\textbf{1-1}} & \multicolumn{1}{c}{\textbf{N-1}} & \multicolumn{1}{c}{\textbf{N-M}} & \multicolumn{1}{c}{\textbf{All}} & \multicolumn{1}{c}{} & \multicolumn{1}{c}{} \\
 \midrule
\textbf{BERT-base} & {1.59} & 15.46 & 10.33 & 9.12 & 67.94 & 32.67 & 23.54 & 30.83 & 14.29 & 15.88 \\
\textbf{BERT-large}  & {1.59} & 15.53 & 12.16 & 9.76 & 74.23 & 31.30 & 25.30 & 31.05 & \underline{17.61} & \textbf{18.72} \\
\textbf{LM-CORE(b,y)}  & \textbf{64.44} & \underline{52.71} & \underline{50.98} & \underline{56.04} & {74.37} & {51.18} & {34.57} & {45.83} & 15.61 & 14.78 \\
\textbf{LM-CORE(b,w)} & 0.66 & {37.62} & {31.11} & {23.13} & \textbf{81.79} & \underline{59.86} & \underline{45.48} & \underline{55.32} & {17.28} & \underline{16.15}  \\
\textbf{REALM}  & \underline{49.06} & \textbf{79.56} & \textbf{64.13} & \textbf{67.36} & \underline{55.81} & \textbf{69.54} & \textbf{66.98} & \textbf{68.18} & \textbf{27.96} & 4.78  \\
\bottomrule
\end{tabular}%
}
\vspace{-1em}
\end{table*}

Note that while REALM is similar to our proposed solution as far as the idea of retrieving relevant knowledge is concerned, the key difference in the two approaches lies in the source of knowledge being used. REALM relies on an unstructured text corpus (Wikipedia) as the source of knowledge and employs a computationally complex retrieve and read paradigm requiring additional training of the knowledge retriever model. Our proposed solution, on the other hand, uses structured knowledge which offers the advantage of being (almost) unambiguous and less resource-hungry compared to unstructured text. We present the resource requirements of our approach and REALM in Table 9. Note that the size of the external knowledge (in number of words) used by REALM is an order of magnitude greater, and requires three times the number of parameters compared to our model. Furthermore, REALM was trained for 200K steps with a batch size of 512 on an 80 TPU cluster, whereas our proposed solution is much more efficient being trained for 1K steps with a batch size of 512 on a machine with 8 Nvidia A100 GPUs. This computational efficiency of our proposed solution allows us to continue further work on improving our performance by enhancing the structured knowledge base and bridge the performance gap with more complex and computationally expensive models such as REALM.
\begin{table}[h]
\label{tab:resources-realm}
\centering
\begin{small}
\begin{tabular}{@{}llll@{}}
\toprule
\multirow{2}{*}{\textbf{Model}} & \textbf{no. of} & \textbf{Retrieval} & \textbf{Resources} \\
& \textbf{params}& \textbf{corpus size} & \textbf{used}\\
\midrule
LM-CORE(b,y)  &     110M    &        17M KB triples          &   8 GPUs             \\
&  & ($\sim$100M words)              &    \\
LM-CORE(b,w)  &     110M    &        21M KB triples          &   8 GPUs             \\
&  & ($\sim$140M words)              &    \\
REALM &      330M    &   5.5M documents                   &   80 TPUs             \\
&   &   ($\sim$2B words)                 &    \\
\bottomrule
\end{tabular}
\caption{Resource requirements of LM-CORE and REALM. REALM requires additional ICT pre-training over all Wikipedia documents for initialization.}
\end{small}
\end{table}

\section{LAMA Evaluation}
\label{sec:lama_eval}
We use the official LAMA data code\footnote{\url{https://github.com/facebookresearch/LAMA}} for evaluating P@1 numbers in Table \ref{table:LAMA}. All the BERT-based models are evaluated using this repository. The LAMA code provides functionality for evaluating RoBERTa models trained in the fairseq framework. Hence, we evaluate RoBERTa-base, RoBERTa-large and KEPLER~\cite{10.1162/tacl_a_00360} using this code. The KEPLER repo also points to this code for evaluation. CoLAKE~\cite{colake}, has adapted the official code\footnote{\url{https://github.com/txsun1997/CoLAKE}} to allow huggingface transformer checkpoints as input, and hence this code is used for CoLAKE, LM-CORE(r,y) and LM-CORE(r,w) evaluation. We ensure the model vocabularies and data is consistent across evaluation. We have used author provided/recommended code and publicly available checkpoints from the official code repositories for all baselines.

\subsection{LAMA: Qualitative Analysis}
\label{sec:lama-qualitative}

Table \ref{table:LAMA_analysis_bert} and \ref{table:LAMA_analysis_roberta} show examples spanning different relationships in LAMA where LM-CORE(b,y) and LM-CORE(r,w) are able to make correct predictions. We also compare the predictions with BERT-base and RoBERTa-base respectively and highlight how these PLMs struggle to make knowledgeable predictions.

\begin{table*}[t]
\centering
\resizebox{\textwidth}{!}{   
\begin{tabular}{llllll}
\toprule
& \multirow{2}{*}{\textbf{Relation}} & \multirow{2}{*}{\textbf{Input query}} & {\textbf{LM-CORE(b,y)}} & \textbf{BERT-base} & \multirow{2}{*}{\textbf{Candidates}} \\
& & & \textbf{prediction} & \textbf{prediction} &  \\\midrule

\multirow{26}{*}{\rotatebox[origin=c]{90}{Google-RE}} & birth-place & \multirow{2}{*}{\begin{tabular}[c]{@{}l@{}}Stanley Corrsin was born \\ in \underline{\hspace{2em}}.\end{tabular}} & Philadelphia & London & (Stanley Corrsin birth date 1920-04-03) \\
& & & & & (Stanley Corrsin nationality United States)\\
& & & & & \textbf{(Stanley Corrsin birth place Philadelphia)}\\
& & & & & (Stanley Corrsin given name Stanley (given name))\\
& & & & & (Stanley Corrsin death date 1986-06-02)\\
& &  & & & (Stanley Corrsin has occupation Physicist)\\
& & & & & (Stanley Corrsin alumni of University of Pennsylvania)\\
& & & & & (Stanley Corrsin member of \\
& & & & &  American Academy of Arts and Sciences)\\

& & & & & \\

& birth-date & \multirow{1}{*}{\begin{tabular}[c]{@{}l@{}}Tom Coppola (born \underline{\hspace{2em}}).\end{tabular}} & 1945 & 1975 & \textbf{(Tom Coppola birth date 1945-06-06)}\\
& & & & & (Tom Coppola given name Tom (given name))\\
& & & & & (Tom Coppola nationality United States)\\
& & & & & (Tom Coppola family name Coppola (surname))\\
& & & & & (Tom Coppola alumni of USC Thornton School of Music)\\
& &  & & & (Christopher Coppola birth date 1962-01-25)\\
& & & & & (Anton Coppola nationality United States)\\
& & & & & (Chris Coppola birth date 1962-01-25)\\

& & & & & \\

& death-place & \multirow{2}{*}{\begin{tabular}[c]{@{}l@{}} Aglaja Orgeni died \\ in \underline{\hspace{2em}}.\end{tabular}} & Vienna & Bucharest & (Aglaja Orgeni death date 1926-03-15) \\
& & & & & \textbf{(Aglaja Orgeni death place Vienna)}\\
& & & & & (Aglaja Orgeni birth date 1841-12-17)\\
& & & & & (Aglaja Orgeni birth place Rimavská Sobota)\\
& & & & & (Aglaja Orgeni nationality Austria)\\
& &  & & & (Aglaja Orgeni nationality Hungary)\\
& & & & & (Aglaja Orgeni has occupation Opera singer)\\
& & & & & (Aglaja Orgeni death place Vienna)\\

\midrule

\multirow{35}{*}{\rotatebox[origin=c]{90}{T-REx}} & P106 & \multirow{2}{*}{\begin{tabular}[c]{@{}l@{}} Cigoli is a \underline{\hspace{2em}}\\ by profession.\end{tabular}} & architect & lawyer & \textbf{(Cigoli has occupation Architect)}\\
& & & & & (Cigoli nationality Italy)\\
& & & & & (Cino Cinelli has occupation Businessperson)\\
& & & & & (Francesco Cirio has occupation Businessperson)\\
& & & & & (Cigoli birth place Cigoli, San Miniato)\\
& & & & & (Emilio Cigoli has occupation Stage actor)\\
& & & & & (Francesco Cigalini has occupation Mathematician)\\
& & & & & (Ciputra has occupation Businessperson)\\

& & & & & \\

& P463 & \multirow{2}{*}{\begin{tabular}[c]{@{}l@{}} Phil Mogg is a \\member of  \underline{\hspace{2em}}.\end{tabular}} & UFO & parliament & \textbf{(Phil Mogg member of UFO (band))} \\
& & & & & (Phil Mogg nationality United Kingdom)\\
& & & & & (Phil Mogg birth date 1948-04-15)\\
& & & & & (Phil Mogg birth place London)\\
& & & & & (Mo Mozzali member of Minneapolis Millers)\\
& &  & & & (John Mogg, Baron Mogg nationality United Kingdom)\\
& & & & & (Jamie Moyer member of Colorado Rockies)\\
& & & & & (Jamie Moyer member of Philadelphia Phillies)\\

& & & & & \\

& P407 & \multirow{2}{*}{\begin{tabular}[c]{@{}l@{}} Summerfolk was written\\ in \underline{\hspace{2em}}.\end{tabular}} & russian & english & (Summerland (novel) in language English language) \\
& & & & & \textbf{(Summerfolk in language Russian language)}\\
& & & & & (The World That Summer genre Neofolk)\\
& & & & & (Summerfolk author Maxim Gorky)\\
& &  & & & (Summer (novel) in language English language)\\
& & & & & (Summertime (novel) in language English language)\\
& & & & & (A Summer Tale date published 2000)\\
& & & & & (Summerteeth in language English language
)\\

& & & & & \\

& P1303 & \multirow{2}{*}{\begin{tabular}[c]{@{}l@{}} Nigel Pulsford plays \\ \underline{\hspace{2em}}.\end{tabular}} & guitar & sgt & \textbf{(Nigel Pulsford has occupation Guitarist)}\\
& & & & & (Nigel Pulsford given name Nigel)\\
& & & & & (Nigel Pulsford birth date 1963-04-11)\\
& & & & & (Nigel Pulsford nationality United Kingdom)\\
& & & & & (Nigel Pulsford nationality Wales)\\
& &  & & & (Nigel Pulsford birth place Newport, Wales)\\
& & & & & (William Pulsford nationality) \\
& & & & & United Kingdom of Great Britain and Ireland)\\
& & & & & (Reginald Purdell has occupation Actor)\\
\bottomrule
\end{tabular}%
}
\caption{Illustrative examples of cases where LM-CORE(b,y) model successfully output the correct completions for various probes in LAMA. Candidates containing correct fact are in bold.}
\label{table:LAMA_analysis_bert}
\end{table*}

\begin{table*}[t]
\centering
\resizebox{\textwidth}{!}{   
\begin{tabular}{llllll}
\toprule
& \multirow{2}{*}{\textbf{Relation}} & \multirow{2}{*}{\textbf{Input query}} & {\textbf{LM-CORE(r,w)}} & \textbf{RoBERTa-base} & \multirow{2}{*}{\textbf{Candidates}} \\
& & & \textbf{prediction} & \textbf{prediction} &  \\\midrule

\multirow{10}{*}{\rotatebox[origin=c]{90}{Google-RE}} & birth-place & \multirow{2}{*}{\begin{tabular}[c]{@{}l@{}}Sebastiano Maffettone \\ was born in \underline{\hspace{2em}}.\end{tabular}} & Naples & Rome & \textbf{(Sebastiano Maffettone place of birth Naples)} \\
& & & & & (Sebastiano Mazzoni place of birth Florence)\\
& & & & & (Sebastiano Mocenigo place of birth Venice)\\
& & & & & (Sebastiano Martinelli place of birth Italy)\\
& & & & & (Sebastiano Baggio place of birth Italy)\\
& & & & & (Stanley Corrsin has occupation Physicist)\\
& & & & & (Sebastiano Vassalli place of birth Genoa)\\
& & & & & (Sebastiano Poma place of birth Parma) \\

\midrule

\multirow{38}{*}{\rotatebox[origin=c]{90}{T-REx}} & P413 & \multirow{2}{*}{\begin{tabular}[c]{@{}l@{}} Rivaldo plays in \underline{\hspace{2em}}\\ position.\end{tabular}} & midfielder & the & (Rivaldo position played on team forward)\\
& & & & & \textbf{(Rivaldo position played on team midfielder)}\\
& & & & & (Rivaldo Gonzalez position played on team midfielder)\\
& & & & & (Rivaldo Coetzee position played on team defender)\\
& & & & & (Rivaldo Vítor Mosca Ferreira Júnior position played on team forward)\\
& &  & & & (Rivaldo member of sports team brazil national football team)\\
& & & & & (Rivaldo member of sports team brazil national under-20 football team)\\
& & & & & (Rivaldo member of sports team São Paulo fc)\\
            
& & & & & \\

& P176 & \multirow{2}{*}{\begin{tabular}[c]{@{}l@{}} Amiibo is \\produced by  \underline{\hspace{2em}}.\end{tabular}} & Nintendo & Samsung & \textbf{(amiibo manufacturer nintendo)} \\
& & & & & (amiibo tap: nintendo's greatest bits publisher nintendo)\\
& & & & & (animal crossing: amiibo festival publisher nintendo)\\
& & & & & (amiibo tap: nintendo's greatest bits platform wii u)\\
& & & & & (animal crossing: amiibo festival developer nintendo entertainment\\
& & & & & planning \& development)\\
& &  & & & (animal crossing: amiibo festival platform wii u)\\
& & & & & (amiibo instance of internet protocol)\\
& & & & & (animal crossing: amiibo festival genre party game)\\

& & & & & \\

& P138 & \multirow{2}{*}{\begin{tabular}[c]{@{}l@{}} Uraninite is named\\ after \underline{\hspace{2em}}.\end{tabular}} & uranium & the &\textbf{(uraninite named after uranium)} \\
& & & & & (uraniborg named after urania)\\
& & & & & (uranopilite named after compound)\\
& & & & & (uranopilite named after uranium)\\
& & & & & (uraniinae instance of taxon)\\
& & & & & (urania parent taxon uraniinae)\\
& & & & & (uranocircite-ii named after uranium)\\
& & & & & (30 urania named after urania) \\

& & & & & \\

& P159 & \multirow{2}{*}{\begin{tabular}[c]{@{}l@{}} The headquarter of\\Stelco is in \underline{\hspace{2em}}.\end{tabular}} & Hamilton & Madrid & \textbf{(Stelco headquarters location Hamilton)}\\
& & & & & (Stelco lake erie works located in the administrative \\
& & & & & territorial entity Ontario)\\
& & & & & (Stelco owned by U.S. steel)\\
& & & & & (Stelco lake erie works country Canada)\\
& & & & & (Stelco industry ferrous metallurgy)\\
& &  & & & (Stelco instance of business)\\
& & & & & (Stec, inc. headquarters location California \\
& & & & & (Stekey located in the administrative territorial entity louisiana)\\

& & & & & \\

& P37 & \multirow{2}{*}{\begin{tabular}[c]{@{}l@{}} The official language\\of Virrat is \underline{\hspace{2em}}.\end{tabular}} & Finnish & English & \textbf{(Virrat official language Finnish)}\\
& & & & & (Virrat country Finland)\\
& & & & & (Virrat located in the administrative territorial entity Pirkanmaa)\\
& & & & & (Virrat located in time zone utc+2)\\
& & & & & (Virrat located in time zone utc+03:00)\\
& & & & & (Virrat instance of municipality of Finland)\\
& & & & & (Virrat instance of town) \\
& & & & & (Virrat instance of city)\\

\bottomrule
\end{tabular}%
}
\caption{Illustrative examples of cases where LM-CORE(r,w) model successfully output the correct completions for various probes in LAMA. Candidates containing correct fact are in bold.}
\label{table:LAMA_analysis_roberta}
\end{table*}

\subsection{LAMA: Error Analysis}
\label{sec:lama-error}
\begin{table*}[t]
\centering
\resizebox{\textwidth}{!}{   
\begin{tabular}{lllll}
\toprule
\textbf{\begin{tabular}[c]{@{}l@{}}Input\\ Query\end{tabular}} & \textbf{\begin{tabular}[c]{@{}l@{}}Expected \\ Answer\end{tabular}} & \textbf{\begin{tabular}[c]{@{}l@{}}Model\\ Output\end{tabular}} & \textbf{\begin{tabular}[c]{@{}l@{}}Retrieved\\ Candidates\end{tabular}} & \textbf{Comments} \\

\midrule
\multirow{1}{*}{\begin{tabular}[c]{@{}l@{}} Hans Gefors was\\ born in \underline{\hspace{2em}}.\end{tabular}} & Stockholm & Hamburg & (Hans Raj Hans birth date 1953-11-30) & \multirow{8}{10em}{Corresponding fact not present in KB. We speculate that the candidate in bold led the model to predict Hamburg. BERT
predicted \textit{Oslo} as the answer.}\\
& & & (Hans Raj Hans given name Hans (name)) & \\ %
& & & (Hans Raj Hans nationality India) & \\ %
& & & (Hans Raj Hans has occupation Politician) & \\ %
& & & (Hans Raj Hans member of Indian National Congress) & \\ %
& &  & (\textbf{Claus Gerson birth place Hamburg}) & \\ %
& & & (Hans Geister birth date 1928-09-28) \\ %
& & & (Hans Gericke nationality Germany) & \\ %

\midrule

\multirow{1}{*}{\begin{tabular}[c]{@{}l@{}} Victor Salvi plays\\ \underline{\hspace{2em}}.\end{tabular}} & harp & quarterback  & (Victor Salvi given name Victor (name)) & \multirow{8}{10em}{Corresponding fact not present in KB} \\
& & & (Victor Salvi nationality United States) & \\ %
& & & (Victor Salvi death place Milan)& \\ %
& & & (Victor Salvi death date 2015-05-10) & \\ %
& & & (Victor Salvi birth place Chicago) & \\ %
& &  & (Victor Salvi birth date 1920-03-04) & \\ %
& & & (Joan Lui actor Francesco Salvi) & \\ %
& & & (Victor Salvi birth date 1920-03-04) & \\ %

\midrule

\multirow{1}{*}{\begin{tabular}[c]{@{}l@{}} CBeebies is owned\\ by \underline{\hspace{2em}}.\end{tabular}} & BBC & Microsoft  & (CBeebies founding date 2002) & \multirow{8}{10em}{Corresponding fact not present in KB.} \\
& & & (Gigglebiz creator CBeebies) & \\ %
& & & (CBEF contained in place Ontario)& \\ %
& & & (CBEF location Ontario) & \\ %
& & & (Bambi production company The Walt Disney Company) & \\ %
& &  & (CBE Software founding date 2006) & \\ %
& & & (Paddington Bear (TV series) production company ITV Central) & \\ %
& & & (CBS Interactive parent organization CBS Corporation) & \\ %

\midrule

\multirow{1}{*}{\begin{tabular}[c]{@{}l@{}} Ivan Petch was\\ born in \underline{\hspace{2em}}.\end{tabular}} & Concord & Sydney & (Ivan Petch birth date 1939-03-01) &  \multirow{8}{10em}{Correct fact is retrieved. However, the model is still not able to predict correct output.}\\
& & & (\textbf{Ivan Petch birth place Concord, New South Wales}) & \\ %
& & & (Ivan Petch family name Petch)& \\ %
& & & (Ivan Petch given name Ivan (name)) & \\ %
& & & (Ivan Petch nationality Australia) & \\ %
& &  & (Ivan Petch has occupation Politician) & \\ %
& & & (Ivan Petch has occupation Electrical engineer) & \\ %
& & & (Ivan Petch alumni of Fort Street High School) & \\ %

\midrule
\multirow{1}{*}{\begin{tabular}[c]{@{}l@{}} Scientist was\\ born in \underline{\hspace{2em}}.\end{tabular}} & Kingston & London  & (Scientist (musician) birth date 1960-04-18) & \multirow{8}{10em}{Ambiguous query, leads to poor retrieval results.}  \\
& & & (Thomas Young (scientist) has occupation Physicist) & \\ %
& & & (I Am a Scientist date published 1994)& \\ %
& & & (Thomas Prince (scientist) has occupation Physicist) & \\ %
& & & (Bambi production company The Walt Disney Company) & \\ %
& &  & (Allen Taylor (scientist) nationality United States) & \\ %
& & & (Lawrence Roberts (scientist) nationality United States) & \\ %
& & & (David Thomas (Canadian scientist) has occupation Biochemist) & \\ %

\midrule
\multirow{1}{*}{\begin{tabular}[c]{@{}l@{}} Moldova shares\\ border with \underline{\hspace{2em}}.\end{tabular}} & Ukraine & Romania  & (Moldova shares border with Ukraine) & \multirow{8}{10em}{Multiple answers correct, however, LAMA considers only one.}  \\
& & & (Moldova shares border with Romania) & \\ %
& & & (Moldova shares border with aa)& \\ %
& & & (Moldova shares border with Jabara) & \\ %
& & & (Moldova Nouã shares border with Bela Crkva) & \\ %
& &  & (Moldova contains administrative territorial entity Transnistria) & \\ %
& & & (Moldova diplomatic relation Russia) & \\ %
& & & (Moldova diplomatic relation European Union) & \\ %

\bottomrule
\end{tabular}%
}
\caption{Illustrative examples of cases where the proposed solution produced incorrect completions.}
\label{table:LAMA_error_analysis}
\end{table*}

We present various failure cases for LM-CORE in Table \ref{table:LAMA_error_analysis}. These are representative of the type of errors we encountered, however, we observed that majority of the errors resulted due to correct facts missing from the KB.  

\subsection{Complete LAMA-UHN results}

The complete LAMA-UHN results over all subsets of Google-RE and T-REx can be found in Table \ref{table:LAMA-uhn-complete}.
\begin{table*}[]
\centering
\caption{Mean precision at one (P@1) of various models on LAMA-UHN probe. Best results are highlighted in bold and the second best performance is underlined}
\label{table:LAMA-uhn-complete}
\begin{tabular}{@{}lccccccccc@{}}
\toprule
 & \multirow{2}{*}{\textbf{Complete}} & \multicolumn{4}{c}{\textbf{Google-RE}} & \multicolumn{4}{c}{\textbf{T-REx}} \\  \cmidrule(l){3-6} \cmidrule(l){7-10} 
 &  & \textbf{DoB} & \textbf{PoB} & \textbf{PoD} & \textbf{All} & \multicolumn{1}{c}{\textbf{1-1}} & \multicolumn{1}{c}{\textbf{N-1}} & \multicolumn{1}{c}{\textbf{N-M}} & \multicolumn{1}{c}{\textbf{All}} \\ 
 \midrule
 \midrule
\multicolumn{10}{@{}l}{\textit{BERT-based models}} \\
\midrule
\textbf{BERT-base} & 18.72 & \underline{1.59} & 6.98 & 3.98 & 4.18 & 62.86 & \multicolumn{1}{c}{21.99} & \multicolumn{1}{c}{17.32} & \multicolumn{1}{c}{22.16} \\
\textbf{BERT-large} & 19.92 & \underline{1.59} & 7.71 & 5.66 & 49.86 & 70.13 & \multicolumn{1}{c}{22.35} & \multicolumn{1}{c}{19.62} & \multicolumn{1}{c}{23.62} \\
\textbf{ERNIE} & 15.81 & 1.42 & 6.57 & 1.38 & 3.12 & 55.68 & \multicolumn{1}{c}{17.90} & \multicolumn{1}{c}{15.40} & \multicolumn{1}{c}{18.76} \\
\textbf{LM-CORE(b,y)} & \underline{41.33} & \textbf{64.44} & \textbf{46.80} & \textbf{46.02} & \textbf{52.42} & \underline{71.04} & \underline{44.42} & \underline{29.11} & \underline{39.75} \\
\textbf{LM-CORE(b,w)} & \textbf{45.50} & 0.66 & \underline{30.93} & \underline{23.39} & \underline{18.33} & \textbf{80.05} & \textbf{55.16} & \textbf{41.17} &\textbf{50.92} \\
 \midrule
 \midrule
\multicolumn{10}{@{}l}{\textit{RoBERTa-based models}} \\
\midrule
\textbf{RoBERTa} & 13.66 & \underline{1.85} & 4.18 & 0.55 & 2.19 & \multicolumn{1}{l}{53.36} & 13.99 & 15.80 & 16.62 \\
\textbf{RoBERTa-large} & 17.99 & 1.41 & 5.68 & 0.36 & 2.48 & \multicolumn{1}{l}{\underline{67.27}} & 20.47 & 17.86 & 21.74 \\
\textbf{KEPLER} & 12.46 & 1.47 & 4.88 & 0.91 & 2.42 & 48.70 & \multicolumn{1}{c}{12.60} & \multicolumn{1}{c}{14.44} & \multicolumn{1}{c}{15.08} \\
\textbf{CoLAKE} & 17.16 & 1.79 & 6.89 & 5.83 & 4.84  & \multicolumn{1}{l}{59.84} & 18.85 & 17.61 & 20.37 \\
\textbf{LM-CORE(r,y)} & \underline{34.25} & \textbf{46.33} & \textbf{35.66} & \underline{19.31} & \textbf{33.77} & 63.83 & \underline{37.74} & \underline{25.20} & \underline{34.12} \\
\textbf{LM-CORE(r,w)} & \textbf{44.75} & 0.38 & \underline{24.80} & \textbf{20.04} & \underline{15.07} & \textbf{67.53} & \multicolumn{1}{c}{\textbf{55.72}} & \multicolumn{1}{c}{\textbf{39.77}} & \multicolumn{1}{c}{\textbf{50.07}} \\ \bottomrule
\end{tabular}
\end{table*}

\section{Downstream Evaluation}
\label{sec:downstream_eval}

We discuss the experimental setup and hyperparameter settings for our downstream tasks.

\subsection{Zero Shot RE}

We consider the open domain version of Zero Short RE~\citep{zsre} from~\citet{kilt}. The dataset is split into three disjoint sets -- train (147,909 samples, 84 relations), dev (3,724 samples, 12 relations) and test (4,966 samples, 24 relations). The systems are evaluated on relations never seen during training.

We fine tune our model for 2 epochs with a batch size of 96. We use the Adam~\cite{DBLP:journals/corr/KingmaB14} optimizer and a learning rate of 3e-5. We performed multiple trials by tuning the number of epochs in \{1, 2, 5\}.

\subsection{Web Questions}

Web Questions~\cite{webqa} was created using questions that were
sampled from the Google Suggest API. We used the same splits as~\citet{orqa} with training, dev and test sets containing 3417, 361 and 2032 samples respectively.

We fine tuned our model for 20 epochs -- we experimented with number of epochs in \{10, 20, 30, 50\}. We use the Adam~\cite{DBLP:journals/corr/KingmaB14} optimizer and a learning rate of 3e-5.

\section{Risks Statement}
This work considers training of large language models using large textual corpora as well as structured knowledge bases. The model learns the nuances of the language and correlations between different real-world entities based on the data that is being used for training the model. Hence, there is a chance that the biases and noise in the training data will creep into the model parameters as well that can lead to a biased model behavior. We need to be careful in deploying the model and extrapolating the output of the model in applications such as search, conversational systems and recommendation systems where model's inherent biases can lead to catastrophic impacts on the user.

\end{document}